\pdfoutput=1
\documentclass[10pt,twocolumn,letterpaper]{article}
\usepackage[pagenumbers]{cvpr} %
\usepackage[accsupp]{axessibility}

\usepackage{amsmath,amsfonts}
\usepackage{algorithmic}
\usepackage{array}
\usepackage[caption=false,font=normalsize,labelfont=sf,textfont=sf]{subfig}
\usepackage{textcomp}
\usepackage{mathtools}
\usepackage{epsfig}
\usepackage{stfloats}
\usepackage{url}
\usepackage{verbatim}
\usepackage{times}
\usepackage{graphicx}
\usepackage{amssymb}

\usepackage[table]{xcolor}
\usepackage{gensymb}
\usepackage{multirow}
\usepackage{color}   
\usepackage{booktabs}
\definecolor{gray}{RGB}{229, 230, 230}
\def\360{360$\degree$}

\begin{document}
\title{CitySeg: A 3D Open Vocabulary Semantic Segmentation\\ Foundation Model in City-scale Scenarios}
\author{
Jialei Xu\thanks{Corresponding author, xujialei5@huawei.com}  ~~~
Zizhuang Wei ~~~
Weikang You ~~~
Linyun Li ~~~
Weijian Sun\\
\\
Huawei Technologies Co., Ltd.
}
\maketitle


\begin{abstract}

Semantic segmentation of city-scale point clouds is a critical technology for Unmanned Aerial Vehicle (UAV) perception systems, enabling the classification of 3D points without relying on any visual information to achieve comprehensive 3D understanding.
However, existing models are frequently constrained by the limited scale of 3D data and the domain gap between datasets, which lead to reduced generalization capability. To address these challenges, we propose CitySeg, a foundation model for city-scale point cloud semantic segmentation that incorporates text modality to achieve open vocabulary segmentation and zero-shot inference.
Specifically, in order to mitigate the issue of non-uniform data distribution across multiple domains, we customize the data preprocessing rules, and propose a local-global cross-attention network to enhance the perception capabilities of point networks in UAV scenarios. To resolve semantic label discrepancies across datasets, we introduce a hierarchical classification strategy. A hierarchical graph established according to the data annotation rules consolidates the data labels, and the graph encoder is used to model the hierarchical relationships between categories. In addition, we propose a two-stage training strategy and employ hinge loss to increase the feature separability of subcategories.
Experimental results demonstrate that the proposed CitySeg achieves state-of-the-art (SOTA) performance on nine closed-set benchmarks, significantly outperforming existing approaches. Moreover, for the first time, CitySeg enables zero-shot generalization in city-scale point cloud scenarios without relying on visual information.
\end{abstract}

\begin{figure}[t!]
\centering
\includegraphics[width=8.5cm]{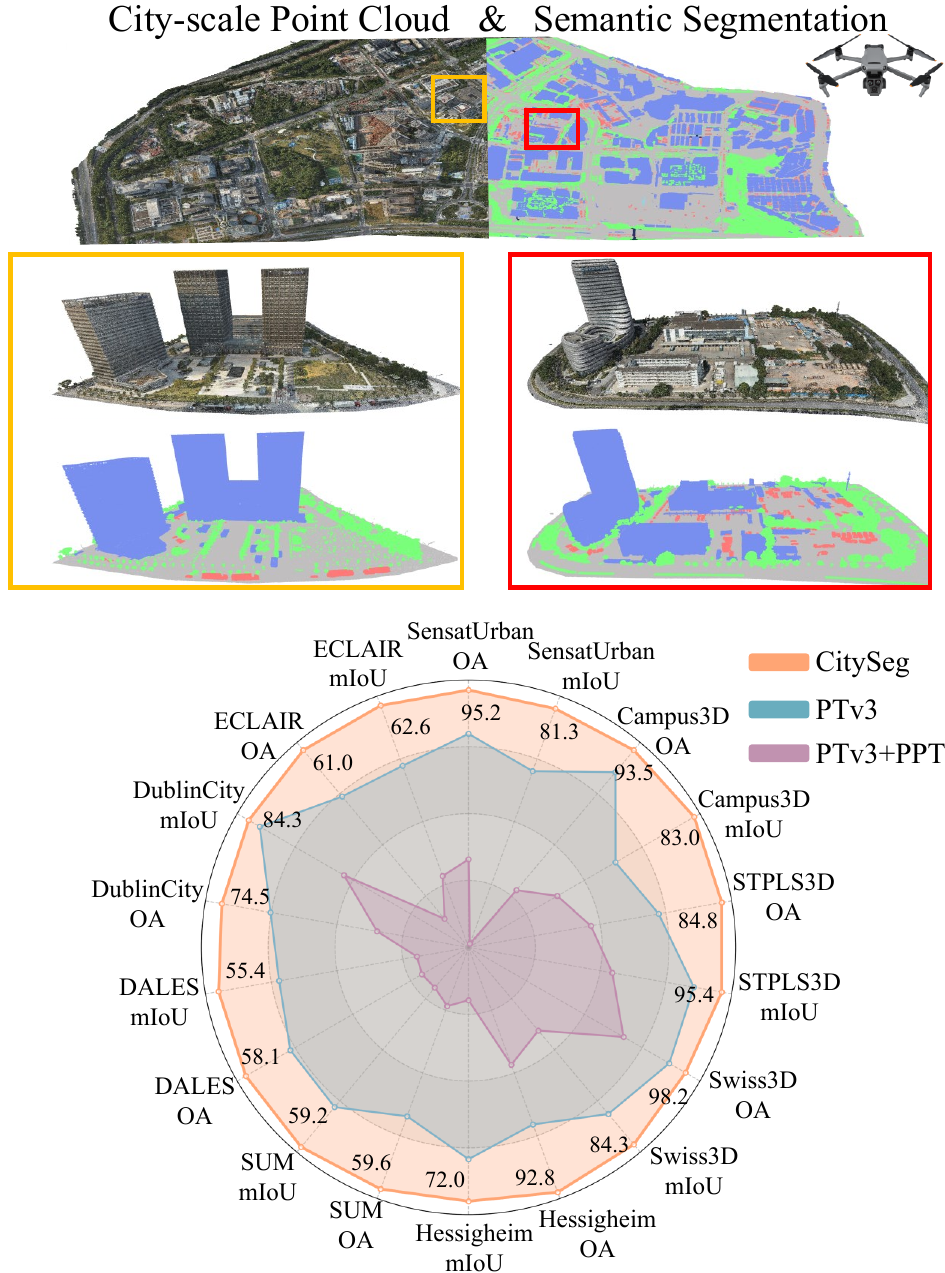}
\caption{\textbf{City-scale point cloud, semantic segmentation results and the radar chart.} The radar chart shows the effectiveness of CitySeg compared to two SOTA methods~\cite{wu2024point,wu2024towards} on 9 benchmarks with two metrics. Abbreviations: mIoU denotes Mean Intersection over Union; OA denotes Overall Accuracy.}
 \label{fig:teaser}
\end{figure}

\section{Introduction}

3D scene understanding~\cite{hou2021exploring,chen2023clip2scene,kim20133d} plays a critical role in Unmanned Aerial Vehicle (UAV) applications~\cite{muchiri2022review,nex2014uav,tsouros2019review}. The main goal is to recognize and classify objects from 3D data, such as point clouds, while minimizing the need for manual annotations. With the rapid development of vision-language foundation models, 3D open vocabulary semantic segmentation is emerging as a mainstream approach, gradually supplanting traditional closed-set semantic segmentation. These open-set methods allow models to assign semantic labels to previously unseen or unlabeled objects in the training data, thus overcoming the limitations of fixed-category approaches.

Existing approaches~\cite{peng2023openscene,jiang2024open,zhu2024open,liu2023weakly,huang2024segment3d} for open vocabulary 3D perception predominantly focus on bridging the gap between 3D and text representations by leveraging 2D images as intermediaries. Typically, these methods either align 3D embeddings with vision-language embedding spaces~\cite{yang2024regionplc,he2024unim} or distill open-vocabulary knowledge from vision-language foundation models into 3D networks~\cite{liu2024segment,xiao20243d}, achieving notable results in indoor environments~\cite{ding2023pla,nguyen2024open3dis,takmaz2025search3d} and self-driving scenarios~\cite{vobecky2023pop,hess2024lidarclip}. 
However, these techniques encounter significant challenges in UAV applications. On the one hand, certain UAV systems lack image sensors and rely solely on LiDARs and positioning systems for data acquisition~\cite{zolanvari2019dublincity,varney2020dales}. On the other hand, for UAVs equipped with image sensors, existing vision-language foundation models often perform suboptimally when processing aerial images. This is mainly because aerial images are rarely present in the training datasets of these foundation models~\cite{liu2024grounding,radford2021learning}. Additionally, aerial images are frequently affected by significant occlusion, further complicating the alignment of 3D point clouds with textual representations.%

A straightforward approach is to adopt the framework of established vision-language foundation models and train a point–text model using a large-scale collection of paired point cloud and text data. However, such research remains limited, primarily due to substantial domain gaps across 3D datasets. Directly combining multiple data sources can result in negative transfer~\cite{wu2024towards},  a phenomenon where discrepancies in data distributions across sources lead to degraded model performance. 
We identify two primary causes for this issue: (1) divergent data distributions, as variations in UAV sensors and data collection algorithms lead to significant differences in the density and scale of 3D point clouds, impeding the generalization capability of current models; and (2) non-uniform 3D data annotations, since different datasets employ varying annotation criteria with different levels of granularity. These challenges motivate us to develop a robust and generalizable foundation model capable of effectively aligning point clouds with textual semantics.

In this paper, we introduce CitySeg, a foundation model for 3D open-vocabulary semantic segmentation in city-scale scenarios, enabling semantic annotation of point clouds through textual descriptions. Specifically, to address distributional disparities across multi-source data, we develop a point network equipped with a local-global cross-attention module, extending the perceptual range in city-scale scenarios and improving robustness to sparse point clouds. Additionally, to address annotation inconsistencies across datasets, we propose a hierarchical classification strategy by constructing a category hierarchy tree based on annotation criteria, thus accommodating varying levels of annotation granularity. 
Finally, we design a two-stage training strategy to enhance the discriminative ability among subcategories and to enable both model fine-tuning and zero-shot inference.
The main contributions of our work are as follows:
\begin{itemize}
\item We introduce CitySeg, a 3D open vocabulary semantic segmentation foundation model tailored for city-scale scenarios, representing the first model to align point clouds with text in UAV applications.
\item We propose a local-global cross-attention module and a hierarchical classification strategy to address the challenges of multi-source data in city-scale scene understanding.
\item Experimental results demonstrate that CitySeg achieves SOTA performance on nine closed-set benchmarks, significantly outperforming existing methods. Furthermore, CitySeg is the first model to achieve zero-shot inference in city-scale scenarios.
\end{itemize}

\section{Related Work}

\noindent\textbf{City-Scale 3D Understanding}.
Low-altitude mobility, particularly through unmanned aerial vehicles (UAVs), has brought transformative advancements to domains such as transportation~\cite{bakirci2024enhancing}, logistics~\cite{jahani2024exploring}, and emergency rescue~\cite{tang2024review}. 
As a key enabling technology for UAVs, city-scale 3D perception has garnered significant attention from the research community~\cite{ren2024k,maxey2024uav,yang2023urbanbis}.
City-scale scenarios are characterized by their large spatial extent and complex lighting conditions, and 3D point clouds serve as the primary representation for such environments.
The acquisition of 3D point cloud data can generally be categorized based on the underlying hardware sensor: LiDAR~\cite{varney2020dales,kolle2021hessigheim} and multi-view stereo (MVS) from optical cameras~\cite{hu2022sensaturban,can2021semantic}. A growing body of research~\cite{wang2022meta,li2023mvpnet,yang2023urbanbis,huang2024openins3d,tang2023all} has explored semantic segmentation of point clouds in city contexts. For example, RandLA-Net~\cite{hu2020randla} introduces an efficient local feature aggregation module that preserves complex local structures by progressively increasing the receptive field for each point. KPConv~\cite{thomas2019kpconv} proposes the Kernel Point Convolution operator, enabling the construction of deep architectures for point cloud classification and segmentation with efficient training and inference. PtB~\cite{du2022push} presents the first end-to-end framework for joint semantic segmentation, boundary detection, and direction prediction in 3D domains. B-Seg~\cite{yang2023urbanbis} advances semantic understanding by classifying city buildings into seven sub-categories.
However, existing approaches to urban 3D understanding are typically trained on individual datasets and lack the zero-shot generalization capability.
In this work, we aim to develop a foundation model for 3D semantic segmentation in city scenarios, delivering both robustness and strong generalization across domains.

\noindent\textbf{Open-Vocabulary 3D Semantic Segmentation.}
In 3D vision, Semantic Abstraction~\cite{ha2022semantic} is the first work to leverage visual-language models for open-world 3D scene understanding, providing significant inspiration for subsequent research. 
Existing approaches to 3D open-vocabulary semantic segmentation can be broadly categorized into two paradigms: model distillation and feature alignment. Model distillation methods transfer knowledge from foundation vision-language models into 3D networks by exploiting the mapping between 2D and 3D data. For example, SEAL~\cite{liu2024segment} is the first to utilize 2D vision foundation models for self-supervised representation learning on large-scale 3D point clouds, enabling the extraction of informative features from automotive point cloud sequences. Liu et al.~\cite{liu2023weakly} distill the open-vocabulary multi-modal knowledge and object reasoning capabilities of CLIP and DINO into a neural radiance field, thereby elevating 2D features to view-consistent 3D segmentation. Xiao et al.~\cite{xiao20243d} present the first method for open-vocabulary 3D panoptic segmentation in autonomous driving scenarios by leveraging large vision-language models.
In contrast, feature alignment approaches aim to directly align 3D representations with vision-language embedding spaces, typically using 2D modalities as intermediaries to bridge 3D-text relationships. OpenScene~\cite{peng2023openscene} predicts dense features for 3D scene points that are co-embedded with text and image pixels in the CLIP feature space. PLA~\cite{ding2023pla} distills knowledge encoded in pre-trained vision-language foundation models through captioning multi-view images from 3D scenes, enabling explicit associations between 3D data and semantically rich textual descriptions. UniM-OV3D~\cite{he2024unim} introduces a unified multimodal network for open-vocabulary 3D scene understanding, establishing dense alignment between modalities to effectively leverage their complementary strengths.

\begin{figure*}[t!]
\centering
\includegraphics[width=18.0cm]{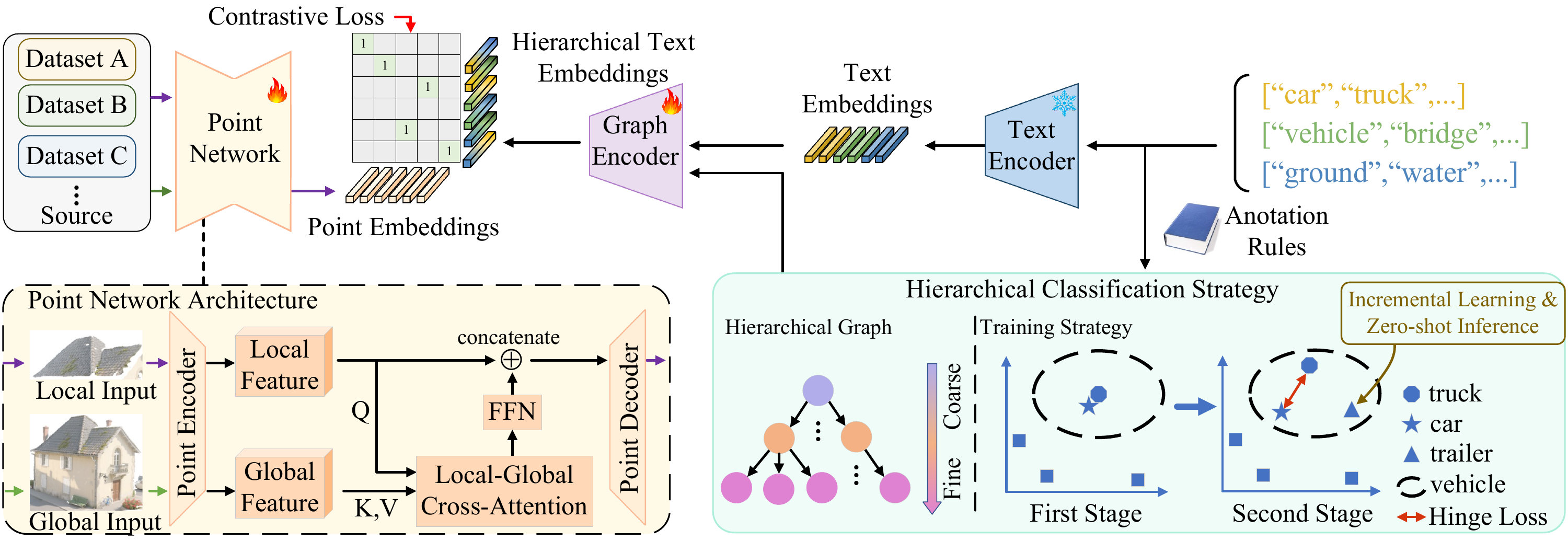}
\caption{\textbf{Overall pipeline of the proposed CitySeg model.} 
Multi-domain 3D point clouds are processed by a customized city-scale point network to generate point embeddings.
A hierarchical classification strategy is applied to the text modality to reconcile differences in semantic granularity across datasets. A graph encoder incorporates category hierarchy into the text embeddings, resulting in hierarchical text embeddings. Both training and inference are performed by computing similarities between point embeddings and hierarchical text embeddings for semantic label prediction.
}
 \label{fig:architecture}
\end{figure*}

\section{Method}
\subsection{Preliminaries}
\noindent\textbf{Task.} Given a point cloud $\mathbf{P} = \{(pos_i,e_i) \mid i=1,...,N\}$ consisting of $N$ points acquired on a UAV platform by LiDARs or multi-view stereo (MVS) algorithms, where $pos_i$  and $e_i$  represent the spatial coordinates and point features (such as intensity, 
elongation, color, $etc.$) of the $i$-th point, the objective is to predict the semantic category of each point and align it with the corresponding textual description.
\noindent\textbf{Data.} We collect and merge city-scale 3D point cloud data from various domains, which present the following challenges for semantic segmentation:
(1) \textbf{Divergent data distributions.} Variations in UAV data collection platforms result in substantial differences in the density and scale of point clouds across datasets. Point cloud density is influenced by LiDAR sensor performance or MVS algorithm design. Such variations reduce the model’s ability to comprehend diverse 3D scenarios and affect the effective receptive field in 3D space, ultimately leading to decreased generalization performance across datasets. Additionally, in city-scale scenarios, the number of points can approach 100 million, which is significantly greater than the typical 10 thousand points in self-driving datasets~\cite{caesar2020nuscenes} and 300 thousand points in indoor datasets~\cite{silberman2012indoor}.  
(2) \textbf{Non-uniform 3D data annotations.} Each dataset employs independent annotation protocols, and identical textual labels may correspond to different object definitions. For instance, in the STPLSD dataset~\cite{chen2022stpls3d}, ``car'' refers to all mobile vehicles, while in the DALES dataset~\cite{varney2020dales}, ``car'' and ``truck'' are treated as separate categories. Directly merging multiple datasets can disrupt model training and induce negative transfer~\cite{wu2024towards}. In the following sections, we detail our strategies to address these challenges.

\subsection{City-scale Point Network}
\noindent\textbf{Data Preprocessing.}
To standardize data formats from multiple domains and unify point cloud density, we design a set of simple yet effective preprocessing strategies. Missing data attributes, such as color information in LiDAR-collected point clouds~\cite{varney2020dales,can2021semantic}, are filled with zero values.
Furthermore, we employ grid sampling~\cite{hu2020randla} to standardize point cloud density, which both unifies 3D spatial granularity and reduces memory consumption.

\noindent\textbf{Point Network.}
Most existing point cloud backbones are designed for self-driving~\cite{hu2020randla,thomas2019kpconv,peng2024oacnns,tian2023cac}, indoor~\cite{silberman2012indoor,dai2017scannet}, or object-level scenarios~\cite{wu20153d,chang2015shapenet}, all operating at scales much smaller than city-scale environments. Due to memory constraints, these methods~\cite{huang2024openins3d,wang2022meta,geng20233dgraphseg} commonly use K-Nearest Neighbors (KNN)~\cite{peterson2009k} to sample subsets of the point cloud, and then combine outputs from multiple regions to obtain full-scene predictions. However, such approaches lack global feature comprehension, which is critical for city-scale scenarios; for example, distinguishing whether a local flat surface corresponds to the ground or a rooftop is challenging without global context. These limitations motivate us to design a local-global feature fusion module to enable holistic understanding at the city scale.

As shown in Fig.~\ref{fig:architecture}, our point network consists of an encoder, a local-global cross-attention module, and a decoder, each based on the Transformer architecture. The \textbf{p}oint \textbf{e}ncoder $f_{pe}$ comprises two input branches: a local branch for fine-grained semantic understanding within high-density, small-scale regions, and a global branch designed to capture broader spatial distributions using sparser point clouds.
Specifically, for each input batch, we apply random sampling~\cite{hu2020randla} to obtain local points $\mathbf{P}_{loc} = \{\mathbf{p}_i \mid i = 1, ..., M\}$, and use KNN to sample a larger set of global points centered on $\mathbf{P}_{loc}$, denoted as $\mathbf{P}_{glo} = \{\mathbf{p}_i \mid i = 1, ..., 10M\}$, where $M$ is the number of local points per batch. Both $\mathbf{P}_{loc}$ and $\mathbf{P}_{glo}$ undergo grid sampling $\mathcal{G}$ to unify density, after which we employ a space-filling curve~\cite{peano1990courbe} to serialize~\cite{wu2024point} each point cloud into a structured format. The point encoder $f_{pe}$ produces local features $\mathbf{F}_{loc}$ and global features $\mathbf{F}_{glo}$ as follows:
\begin{align}
    \mathbf{F}_{loc} &= f_{pe}(\mathcal{G}_{loc}(\mathbf{P}_{loc})), \\\mathbf{F}_{glo} &= f_{pe}(\mathcal{G}_{glo}(\mathbf{P}_{glo})),
\end{align}
where the sizes of the grid sampling $\mathcal{G}_{loc}$ and $\mathcal{G}_{glo}$ are set to 0.2 meters and 1 meter, respectively.

\noindent\textbf{Local-global Cross-attention.}
To enable the local branch to leverage global scene context for enhanced fine-grained semantic discrimination, we propose a local-global cross-attention module. In this module, the local features serve as the query, while the geometric and color features from the global scene are used as the key and value to compute contextualized representations.
Given $\mathbf{F}_{loc}$ and $\mathbf{F}_{glo}$ as input, the attention feature is defined as:
\begin{align}
\mathbf{F}_{attn} = softmax({{Q_{loc}}{K_{glo}}^{T})V_{glo}},
\end{align}
where $Q_{loc}$ is the query derived from the local feature $\mathbf{F}_{loc}$, $K_{glo}$ and $V_{glo}$ are the key and value derived from the global feature $\mathbf{F}_{glo}$, and $softmax$ denotes the softmax function.
The attention feature is subsequently processed by a Feed-Forward Network ($FFN$). The output of $FFN$ is then concatenated ($cat(\cdot,\cdot)$) with the original local features along the channel dimension and passed to the \textbf{p}oint \textbf{d}ecoder $f_{pd}$. The output of the point decoder consists of point embeddings, which are later used for alignment with textual features. The point embeddings $\mathbf{E}_p$ are formulated as follows:
\begin{align}
    \mathbf{E}_p = f_{pd}(cat(\mathbf{F}_{loc}, FFN(\mathbf{ F}_{attn}))) \in \mathbb{R}^{M \times C},
\end{align}
where $C$ is the embedding dimension.
Notably, the point embeddings are computed only for the local input points. The point decoder does not generate embeddings for the global point cloud.  The global features are used exclusively to provide contextual information, thereby assisting local fine-grained semantic segmentation. By integrating this cross-attention mechanism, our customized point network effectively addresses challenges related to varying point cloud density and scale across datasets.

\subsection{Hierarchical Classification Strategy}
\noindent\textbf{Hierarchical Graph.}
Although existing 3D point cloud datasets exhibit annotation inconsistencies, we observe that they all label the fundamental categories present in city-scale scenes.  Annotation conflicts and semantic differences mainly arise from variations in semantic granularity. To address this problem, we propose a hierarchical classification strategy.
We organize labels based to the dataset annotation rules and assign their hierarchy according to semantic granularity. Labels with the broadest coverage and no overlapping objects are defined as base classes, while labels with finer granularity, which are subsets of base classes, are defined as subclasses.
Notably, the texts of labels remain unchanged; the granularity of each label is solely determined by dataset annotation rules, not by textual content.
Formally, we define a hierarchical graph $H = (V, E)$ as a tree structure with nodes $V$ and edges $E$. Each node $v \in V$ represents a label, and its parent node and child nodes are corresponding to its superclass and subclass, respectively.
The root node represents the universal semantic class encompassing all objects.  Each node has exactly one parent node and a unique path from the root node. The tree’s depth is not manually constrained, but is determined by the data annotation rules.

Following previous works~\cite{liu2024multi, ding2023pla, liang2023open}, we utilize a pre-trained \textbf{t}ext \textbf{e}ncoder $f_{te}$ to extract the text embeddings $E_t$ for each label, and perform semantic classification by comparing the similarity between text embeddings and point embeddings.
In our hierarchical graph, every node except the root is associated with a text embedding.
To integrate hierarchical relationships into text embeddings $E_t$, we adopt a \textbf{g}raph \textbf{e}ncoder $f_{ge}$ ~\cite{zhang2021hierarchical} to employ the message-passing mechanism~\cite{wu2020comprehensive}, where each node aggregates information from its neighboring nodes to update its embedding. Multiple layers of neighborhood aggregation can be stacked, facilitating recursive message passing throughout the graph. The updated node embeddings are referred to as hierarchical text embeddings $\hat{E}_t$. Formally, in the $l$-th layer of the graph, the hierarchical text embedding of node $v$, denoted by $\hat{E}_{t,v}^l$ is updated based on the initial text embedding $E_{t,v}$ and the embeddings of its neighbors as follows:
\begin{align}
    \hat{E}_{t,v}^l = f_{ge}(E_{t,v},l,\{E_{t,u}:u\in \mathcal{N}_v \}),
\end{align}
where $\mathcal{N}_v$ denotes the neighboring nodes of $v$. 
The hierarchical text embeddings $\hat{E}_{t}$ encode not only the semantic content of the original text embeddings $\hat{E}_{t}$, but also capture the corresponding granularity and relationships between adjacent nodes in the hierarchical graph.
With this strategy, even when identical textual labels correspond to objects of different granularity across datasets, the hierarchical graph establishes the appropriate label level and relationships according to annotation rules.
Finally, each label is represented by a unique hierarchical text embedding $\hat{E}_{t}$, generated by the graph encoder based on its text and annotation granularity.

\noindent\textbf{Training Strategy.}
To accelerate model convergence and reduce prediction error, we design a two-stage training strategy that simplifies the overall optimization process.
In the first stage, the model is trained to capture the coarse-grained semantic structure of city-scale scenarios. 
Based on the hierarchical graph, all labels are merged into five base categories during initial training phase.
Following previous work~\cite{li2023scaling}, we determine the predicted category by computing the similarity between hierarchical text embeddings and point embeddings. Formally, for the $i$-th point $\mathbf{p}_i$, with corresponding point embedding $E_{p,i}$, class $y_i$, and hierarchical text embedding $\hat{E}_{t,i}^l$, the class probability is given by:
\begin{align}
    prob(y_i | \mathbf{p}_i) = \frac{exp(\sigma(E_{p,i},\hat{E}_{t,i}^l) /\tau)}{\sum_{j=1}^{|y|}exp(\sigma(E_{p,i},\hat{E_{t,j}})/\tau)},
    \label{eq:prob}
\end{align}
where $\tau$ is the temperature coefficient~\cite{radford2021learning} and $\sigma(\cdot,\cdot)$ denotes the cosine similarity function.
We adopt Cross Entropy loss~\cite{oord2018representation}, $L_{CE}$, as the alignment criterion.
After this stage, the model acquires a foundational semantic understanding of the basic categories, forming a basis for learning finer distinctions.

In the second stage, we further refine the model for fine-grained recognition of all subclasses, especially those that are semantically similar and are previously grouped together during the first stage.
To better separate such closely related categories, we introduce a hinge loss~\cite{rosasco2004loss} to increase the feature separation between sibling nodes in the hierarchical tree.
Specifically, given the point embedding $E_{p,i}$ of the $i$-th point $\mathbf{p}_i$, its corresponding node $v_i$ in the tree, and hierarchical text embedding $\hat{E}_{t,i}^l$, we identify the set $\mathcal{S}$ of sibling nodes of $v_i$, along with their hierarchical text embeddings $\{\hat{E}_{t,j}^l\}_{j=1}^{|\mathcal{S}|}$. The hinge loss is defined as follows:
\begin{equation}
\begin{split}
L_{h}=dist(E_{p,i},\hat{E}_{t,i}^l)+\sum_{j=1}^{|\mathcal{S}|}\max\left(m-dist(E_{p,i},\hat{E}_{t,j}^l),0 \right),
\label{eq:total label}
\end{split}
\end{equation}
where $m$ is the margin constant, and $dist(\cdot,\cdot)$ denotes the Euclidean distance function.
In $L_h$, the first term minimizes the distance between the point embedding $E_{p,i}$ and its ground-truth hierarchical text embedding $\hat{E}_{t,i}^l$, promoting closer alignment. 
The second term ensures that the point embedding ${E}_{p,i}$ is sufficiently distant from the hierarchical text embeddings of its sibling nodes $\{\hat{E}_{t,j}^l\}_{j=1}^{|\mathcal{S}|}$.
The margin $m$ enforces an upper bound on the inter-class distance among sibling nodes, ensuring that the child nodes do not exceed the spatial range defined by the parent node.
The hinge loss $L_{h}$ encourages the graph encoder to distinguish sibling categories,  which are prone to misclassification due to similar representations.
During the second stage, the total loss is:
\begin{equation}
    L = L_{CE} + \alpha L_{h},
\end{equation}
where $\alpha$ is a weight parameter.
The proposed hierarchical graph and two-stage training strategy effectively address annotation discrepancies across multiple datasets and enable robust semantic segmentation in complex, multi-domain scenarios.

\subsection{Incremental Learning and Zero-shot}
CitySeg supports both zero-shot inference and fine-tuning on novel datasets as a point cloud-language foundation model. This subsection details the procedures for fine-tuning CitySeg and performing open-vocabulary segmentation on unseen data.
To enable learning of new categories in novel datasets while mitigating catastrophic forgetting of previously acquired knowledge, we adopt a replay-based incremental learning scheme~\cite{luo2023class,liu2020mnemonics,lopez2017gradient}.
The hierarchical classification framework of CitySeg enables efficient extension to new categories by simply adding nodes to the hierarchical graph.
Specifically, when a new point $\mathbf{p}_k$ and its associated category $y_k$ are encountered, a new leaf node is inserted into the hierarchical tree according to the annotation rules for the new label.
The graph encoder subsequently refines the text embedding of the new node, yielding the corresponding hierarchical text embedding. The classification probability for the newly introduced label is then calculated using Equation~\ref{eq:prob}, and the loss function defined in Equation~\ref{eq:total label} is applied. Zero-shot inference is implemented analogously to the fine-tuning process, except that parameter updates are disabled and no gradients are propagated.

\begin{table*}[t!]
\begin{center}
\renewcommand\arraystretch{1.3}
\setlength\tabcolsep{2.1pt} 
\caption{\protect\textbf{\protect\label{tab:multi-datasets} Quantitative results on the 9 closed-set benchmarks.} \colorbox{gray!55}{Methods trained from scratch on each dataset} are marked in grey, which overfit the benchmarks.  \colorbox{blue!10}{Robust methods trained with multiple datasets} are marked in blue, which are tested with the same weights across all datasets. The \textbf{f}oundational \textbf{m}odel CitySeg-FM  is trained from multiple datasets and inference with the same weight, and further \textbf{f}ine-\textbf{t}uning (FT) helps CitySeg surpass all previous methods. 
}
\small
\begin{tabular}{l|cc|cc|cc|cc|cc|cc|cc|cc|ccccccc}
  \hline
\multirow{2}{*}{\textbf{Methods}}  &
\multicolumn{2}{c|}{SensatUrban} & \multicolumn{2}{c|}{Campus3D}  & \multicolumn{2}{c|}{STPLS3D} &\multicolumn{2}{c|}{Swiss3D}  &\multicolumn{2}{c|}{Hessigheim}  & \multicolumn{2}{c|}{SUM}  & \multicolumn{2}{c|}{DALES}  & \multicolumn{2}{c|}{DublinCity}  & \multicolumn{2}{c}{ECLAIR} \\ 
\cline{2-19}
& OA & mIoU & OA &mIoU & OA&mIoU & OA&mIoU & OA&mIoU & OA&mIoU & OA&mIoU & OA&mIoU & OA & mIoU\\
\midrule 
\rowcolor{gray!55}PointNet++~\cite{qi2017pointnet++} &84.3 &32.9& 62.3 & 47.6& 86.5 & 46.5 &72.9 &45.2& 75.8 & 45.7 &85.5&49.5&95.7&68.3 & 69.1 & 48.7 & 82.5 & 52.0\\
\rowcolor{gray!55}RandLA-Net~\cite{hu2020randla} & 89.8 & 52.7 &72.4 & 53.8 & 89.8&50.5 & 77.1 & 49.6 & 77.2 & 47.4 &74.9&38.6 & 93.1 & 76.4  & 75.7 & 53.1 & 86.8 & 53.7\\
\rowcolor{gray!55}KPConv~\cite{thomas2019kpconv} & 93.2 & 57.6 & 73.1 & 52.9&89.8& 53.7 & 75.2 & 51.3& 78.6& 48.4 &93.3&68.8 &97.8 & 81.1 & 80.1 & 56.2 & 84.7 & 57.0\\
\rowcolor{gray!55}OctFormer~\cite{wang2023octformer} & 88.2 & 60.1 & 76.4 & 54.2 & 90.8 & 54.8 & 75.4 & 50.9 & 76.2 & 50.2 & 90.9 & 68.9 & 95.1 & 82.9 & 77.2 & 57.4 & 85.3 & 54.7 \\
\rowcolor{gray!55}SphereFormer~\cite{lai2023spherical} & 91.7 & 64.5 & 76.3 & 56.0 & 88.4 & 53.5 & 77.5& 53.2 & 80.5 &49.7 & 90.3 & 70.0 & 92.5 & 79.1 & 80.7 & 58.4 & 85.9 & 56.9 \\
\rowcolor{gray!55}OA-CNNs~\cite{peng2024oacnns} & 93.5 & 61.4 & 77.0 & 56.2 &  87.3 & 49.3 & 75.9 & 51.9 & 79.4 &50.4 & 92.4 & 69.3 & 95.8 & 81.5 & 79.4 & 54.9  & 82.4 & 56.2 \\
\rowcolor{gray!55}PTv3~\cite{wu2024point}  & \underline{93.9} & \underline{68.4} & 77.1& 55.8 & \underline{91.4} & 56.4 & 77.8 & 55.6 & 80.8 & 51.6 & \underline{93.5} & 71.2 & \underline{96.4} & \underline{83.2} & 82.1 &  56.9 & \underline{87.4} & 58.8 \\
\midrule
\rowcolor{blue!10}PTv3~+~PPT~\cite{wu2024towards} & 70.3 & 45.8 & 41.2& 38.7 & 64.3 & 35.7 & 63.7 & 36.1 & 67.2 & 35.7 & 63.7 & 53.8 & 73.6 & 63.1 & 57.5 & 37.2 & 57.4 & 42.7 \\
\rowcolor{blue!10}Sonata~\cite{wu2025sonata} & 66.3 & 44.1 & 46.3 & 40.5 & 66.3 & 39.4 & 64.1 & 38.2 &63.1 & 33.8 & 70.3 & 61.2 &78.3 & 67.7 & 61.4 & 44.2 & 56.4 & 40.3 \\ 
\rowcolor{blue!10}CitySeg-FM & 93.6 & 67.8 & \underline{78.2} & \underline{57.3}& 89.3 & \underline{56.5} & \underline{78.9} & \underline{56.0} & \underline{81.2} & \underline{51.9} & 91.7 & \underline{71.7} & 93.5& 80.7 & \underline{83.6} & \underline{59.5} & 87.1 & \underline{59.3} \\
\midrule
CitySeg-FT&\bf95.2 & \bf72.0 & \bf81.3 & \bf59.6 & \bf93.5 & \bf59.2 & \bf83.0& \bf58.1 & \bf84.8 & \bf55.4 & \bf95.4 & \bf74.5 & \bf98.2 & \bf84.3 & \bf84.3 & \bf61.0 & \bf92.8 & \bf62.6 \\
\bottomrule
\end{tabular}
\end{center}
\end{table*}

\section{Experiment}
\subsection{Implementation Details}
CitySeg is implemented using the Pointcept~\cite{pointcept2023} toolbox. We adopt the pre-trained PTv3~\cite{wu2024point} as the point cloud backbone and utilize the open-source CLIP~\cite{radford2021learning} text encoder with pre-trained weights for encoding input text. The text encoder is kept frozen throughout all experiments.
Model training is conducted on 8 NVIDIA A800 GPUs with a total batch size of 16. CitySeg is trained for 100 epochs with a learning rate of 0.005 using the AdamW~\cite{loshchilov2017decoupled} optimizer, with learning rate scheduling performed via the OneCycleLR policy. Only 3D coordinate and color information are used during both training and testing.
For each training batch, the number of sampled points is set to $M$=65536 for the local input branch, and 655360 for the global input branch. We set $\alpha=0.3$ to balance the weight of $L_{CE}$ and $L_{h}$. 
Data augmentation and other hyperparameter settings follow those established for PTv3~\cite{wu2024point}. For test-time augmentation, we adopt the voting strategy as in RandLA-Net~\cite{hu2020randla}.

\subsection{Benchmark Setting}
CitySeg is jointly trained on 9 city-scale point cloud datasets:  SensatUrban~\cite{hu2022sensaturban}, Campus3D~\cite{li2020campus3d}, UrbanBis~\cite{yang2023urbanbis}, STPLS3D~\cite{chen2022stpls3d}, Swiss3Dcities~\cite{can2021semantic}, Hessigheim 3D~\cite{kolle2021hessigheim}, SUM~\cite{gao2021sum}, DALES~\cite{varney2020dales}, DUBLINCITY~\cite{zolanvari2019dublincity}, and ECLAIR~\cite{melekhov2024eclair}.
To thoroughly assess CitySeg’s capabilities, we establish both closed-set and open-set benchmarks, depending on the domain of the test set.

\noindent\textbf{Closed-set Benchmark.}
For closed-set semantic segmentation, we evaluate performance on the nine test sets from the same domains as the training datasets.
CitySeg is compared against seven state-of-the-art point cloud semantic segmentation models: PointNet++~\cite{qi2017pointnet++}, Randla-Net~\cite{hu2020randla}, KPConv~\cite{thomas2019kpconv}, OctFormer~\cite{wang2023octformer}, SphereFormer~\cite{lai2023spherical}, OA-CNNs~\cite{peng2024oacnns}, and PTv3~\cite{wu2024point}. 
As these models lack the capability for cross-domain training, each is trained from scratch on each individual dataset. For the joint training of multiple datasets, we further compare CitySeg to PPT~\cite{wu2024towards} and Sonata~\cite{wu2025sonata}, both based on the PTv3 backbone~\cite{wu2024point}. In this setting, CitySeg is evaluated as a \textbf{f}oundational \textbf{m}odel (CitySeg-FM), tested across multiple datasets with shared weights. For a fair comparison, we also report results for CitySeg after \textbf{f}ine-\textbf{t}uning on each dataset (CitySeg-FT).

\noindent\textbf{Open-set Benchmark.}
For open-set evaluation, the UrbanBIS~\cite{yang2023urbanbis} dataset is excluded from model training and used as an unseen test domain. For traditional closed-set segmentation methods, only results for labels present in the training data are reported. Both PTv3+PPT and CitySeg are jointly trained on the other nine datasets and evaluated on the UrbanBIS dataset.

\noindent\textbf{Evaluation Metrics}. Following previous work~\cite{hu2022sensaturban}, the mean class Intersection-over-Union (mIoU), Overall Accuracy (OA) of all classes, and per-class IoU scores are used as the standard metrics.

\begin{table*}[htbp]
\caption{\textbf{Quantitative results on the open-set benchmark.}
UrbanBIS~\cite{yang2023urbanbis} is used as the open-set test set and is excluded from the training data.
\colorbox{red!10}{Existing SOTA methods trained from scratch} are trained on the Sensaturban dataset~\cite{hu2022sensaturban}, and infer only the labels present in the training set. 
\colorbox{brown!15}{
PTv3 + PPT and the \textbf{f}oundational \textbf{m}odel CitySeg (FM) are
jointly trained on 9 datasets.}
For reference, CitySeg trained on the UrbanBIS training set serves as the upper bound.
}
\small
\renewcommand\arraystretch{1.3}
\setlength\tabcolsep{4.2pt} 
{
\begin{tabular}{l|cc|cc|ccccccc}
    \toprule  
    Methods
    &Training Set & Testing Method
    & \rotatebox{0}{OA}
    & \rotatebox{0}{mIoU}
    & \rotatebox{0}{Terrain}
    & \rotatebox{0}{Vegetation}
    & \rotatebox{0}{Water} 
    & \rotatebox{0}{Bridge} 
    & \rotatebox{0}{Vehicle} 
    & \rotatebox{0}{Boat} 
    & \rotatebox{0}{Building} \\

    \midrule
\rowcolor{red!10}    RandLA-Net~\cite{hu2020randla} & SensatUrban & Zero-Shot  & - & - & 35.4 & 39.1  & 12.5 & 8.4 & 33.2 & - & 58.6  \\
\rowcolor{red!10}    OA-CNNs~\cite{peng2024oacnns} & SensatUrban & Zero-Shot   & - & - & 29.4 &36.2 &  5.1 & 10.4 & 34.8 & - & 58.3\\
\rowcolor{red!10}    PTv3~\cite{zhao2023large} &SensatUrban & Zero-Shot  & -  & - & 54.3 & 46.2 & 24.5 & 18.1 & 24.1 & - & 60.1 \\    
    \midrule
\rowcolor{brown!15}    PTv3 + PPT~\cite{wu2024towards} & 9 Datasets& Zero-Shot   & 53.7 & 42.6 & 36.3 & 80.4 & 10.5 & 15.4 & 54.7& 7.3 & 93.8\\
\rowcolor{brown!15}    \textbf{CitySeg-FM} & 9 Datasets & Zero-Shot   & \underline{89.7} & \underline{65.1} & \underline{81.3} & \underline{89.7} & \underline{60.2} & \underline{23.6} & \underline{73.1} & \underline{32.3} & \underline{96.9} \\
    \midrule
    \textbf{CitySeg} & UrbanBIS & Fully supervised & \bf92.0 & \bf70.4 & \bf82.9 & \bf91.6 & \bf73.2 & \bf27.9 & \bf74.6 & \bf46.4 & \bf97.8\\
    \bottomrule
\end{tabular}
}
\label{tab:zero-shot}
\end{table*}

\begin{table*}[t!]
\caption{{Performance comparisons with existing SOTA methods in each label on SensatUrban\cite{hu2022sensaturban} test set.} Overall Accuracy (OA), mean IoU (mIoU), and per-class IoU scores are reported in the table. ``Tra. road'' denotes the traffic road and ``Str. Fur.'' denotes the street furniture.}
\centering
\small
\renewcommand\arraystretch{1.3}
\setlength\tabcolsep{3pt} 
{
\begin{tabular}{l|lllllllllllllll}
    \toprule  
    Methods
    & \rotatebox{55}{OA(\%)}
    & \rotatebox{55}{mIoU(\%)}
    & \rotatebox{55}{ground}
    & \rotatebox{55}{vegetation}
    & \rotatebox{55}{buildings} 
    & \rotatebox{55}{walls} 
    & \rotatebox{55}{bridge} 
    & \rotatebox{55}{parking} 
    & \rotatebox{55}{rail} 
    & \rotatebox{55}{Tra. road} 
    & \rotatebox{55}{Str. Fur.} 
    & \rotatebox{55}{cars} 
    & \rotatebox{55}{footpath} 
    & \rotatebox{55}{bikes} 
    & \rotatebox{55}{water} \\
    \midrule
    PointNet~\cite{qi2017pointnet}  &80.8 &23.7 &68.0 &89.5 &80.0 &0.0 &0.0 &4.0 &0.0 &31.6 &0.0 &35.1 &0.0 &0.0 &0.0 \\
    PointNet++~\cite{qi2017pointnet++} &84.3 &32.9 &72.5 &94.2 &84.8 &2.7 &2.1 &25.8 &0.0 &31.5 &11.4 &38.8 &7.1 &0.0 &56.9 \\
    TragenConv~\cite{tatarchenko2018tangent}  &77.0 &33.3 &71.5 &91.4 &75.9 &35.2 &0.0 &45.3 &0.0 &26.7 &19.2 &67.6 &0.0 &0.0 &0.0 \\
    SPGraph~\cite{landrieu2018large}  &85.3 &37.3 &69.9 &94.6 &88.9 &32.8 &12.6 &15.8 &15.5 &30.6 &23.0 &56.4 &0.5 &0.0 &44.2 \\
    KPConv~\cite{thomas2019kpconv}  &93.2 &57.6 &\underline{87.1} &\underline{98.9} &95.3 &{74.4} &28.7 &41.4 &0.0 &56.0 &{54.4} &{85.7} &40.4 &0.0 &\underline{86.3}\\
    RandLA~\cite{hu2020randla}  &89.8 &52.7 &80.1 &98.1 &91.6 &48.9 &40.8 &{51.6} &0.0 &56.7 &33.2 &80.1 &32.6 &0.0 &71.3 \\
    LCPFormer~\cite{huang2023lcpformer} &{93.5} &{63.4} &86.5 &98.3 &\underline{96.0} &55.8 &{57.0} &50.6 &{46.3} &{61.4} &51.5 &85.2 &\underline{49.2} &0.0 &86.2\\
    MVPNet~\cite{luo2022mvp} & 93.3& 59.4& 85.1& 98.5 & 95.9 & 66.6 & 57.5 & 52.7 & 0.0 & 61.9 & 49.7 & 81.8 & 43.9 & 0.0 & 78.2\\
    PVCFormer-SA~\cite{zhang2024point} & 93.8 & 62.4 & 83.5 & 97.5 & 94.4 & 68.3 & 51.6 & 51.7 & 50.2 & 58.4 & 48.6 & 79.8 & 47.6 & 0.0 & 79.4\\
    DG-Net~\cite{liu2024semantic} & 92.1& 59.8& 84.9 &98.2 &92.8& 57.0 &50.9& 57.5 &28.1& 61.2 &43.8 &81.7& 42.1& 0.2& 79.4\\
    LG-Net~\cite{zhao2023large} & 90.1& 53.2& 82.2& 98.1& 91.2& 57.7& 47.7& 44.6& 3.3 &51.1& 36.7& 78.7& 34.2& 0.0& 66.1\\
    PTv3~\cite{wu2024point} & \underline{93.9}& \underline{68.4} & 84.6 & 98.5 & 94.2 & \underline{78.4} & \underline{57.9} & \underline{62.7} & \underline{59.4} & \underline{66.3} & \underline{55.1} & \underline{86.4} & 48.5& \underline{12.6} & 84.2 \\
    \midrule
    \textbf{CitySeg} & \bf95.2 & \bf72.0 & \bf 88.4 & \bf99.1 & \bf96.8 & \bf83.5 & \bf64.3 & \bf65.2 & \bf66.7 & \bf69.7 & \bf59.4& \bf89.0& \textbf{49.4}& \bf18.4 & \textbf{86.4}\\
    \bottomrule
\end{tabular}
}
\label{tab:sensaturban}
\end{table*}

\begin{table}[t]
\caption{ \textbf{Ablation study on SensatUrban test set}. LG.Attn. denotes the Local-global cross-attention module, Hier. Class denotes the hierarchical classification strategy, and $L_h$ denotes the Hinge Loss function.}
\label{tab:ablation}
\small
\centering
\renewcommand\arraystretch{1}
\setlength\tabcolsep{3pt} 
{
\begin{tabular}{l|ccc|ccc}
    \toprule  
    Item & LG.Attn. & Hier. Class & $L_{h}$
    & \rotatebox{0}{OA(\%)}
    & \rotatebox{0}{mIoU(\%)} \\
    \midrule
      w/o LG.Attn.&  & \checkmark & \checkmark & 89.8 & 65.0 \\
      w/o Hier. Class& \checkmark & &  \checkmark & 71.9 & 39.4 \\
      w/o $L_h$ & \checkmark & \checkmark & & 91.4 & 65.3 \\
    CitySeg &\checkmark & \checkmark & \checkmark & \bf93.6 & \bf67.8\\
    \bottomrule
\end{tabular}
}
\end{table}

\subsection{Results Comparison}

\noindent\textbf{Performance on the Closed-set Benchmark}. 
In Tab.~\ref{tab:multi-datasets}, we compare the proposed CitySeg with nine existing SOTA models across 9 closed-set city-scale point cloud benchmarks.  
Models trained from scratch remain highly competitive on each benchmark. For example, PTv3 achieves high accuracy on some benchmarks. 
This behavior can largely be attributed to overfitting on each individual dataset, leading to limited domain adaptation capabilities. 
In the setting of joint training of multiple datasets, PTv3+PPT and Sonata exhibit substantially lower overall performance, underscoring the significant domain gap between different city-scale point cloud datasets. These results highlight the difficulty of building a unified model with existing approaches that do not explicitly address inter-domain discrepancies.
In contrast, our foundation model (CitySeg-FM) using the same shared weights for all datasets, consistently outperforms domain-specific models on five out of nine benchmarks and demonstrates superior robustness. Furthermore, additional fine-tuning on each dataset (CitySeg-FT) enables CitySeg to achieve the best performance on all benchmarks, surpassing all existing methods by a substantial margin.

\noindent\textbf{Performance on the Open-set Benchmark}.
Tab.~\ref{tab:zero-shot} summarizes the zero-shot inference performance of models trained from scratch on individual datasets, as well as those jointly trained on multiple datasets. The three closed-set models trained from scratch exhibit poor performance on the open-set benchmark, even for the five categories present in their training data. This result underscores that overfitting to individual datasets, while beneficial in closed-set scenarios (as shown in  Tab.~\ref{tab:multi-datasets}), severely limits generalization.
Similarly, PTv3+PPT, jointly trained on multiple datasets, exhibits poor generalization. The substantial domain gaps among datasets result in degraded performance when directly merging them for training, consistent with trends observed in the closed-set evaluation.
In contrast, our proposed CitySeg model demonstrates outstanding zero-shot generalization. In open-set testing, the foundation model (CitySeg-FM) achieves an overall accuracy (OA) of 89.7, just 2.3 points below the upper bound set by fully supervised CitySeg. This highlights the excellent zero-shot inference ability and robustness of our approach.

\noindent\textbf{Comparison of each label.}
As shown in Tab.~\ref{tab:sensaturban}, we conduct a comprehensive category-level analysis of various methods on the SensatUrban dataset.
Our proposed CitySeg maintains robust performance across all semantic categories.
Compared with the existing SOTA method PTv3, CitySeg achieves significant improvements of 1.3 percentage points (from 93.9 to 95.2) in OA and 3.6 percentage points (from 68.4 to 72.0) in mIoU, unequivocally demonstrating its technical superiority.

\begin{figure*}[t!]
\centering
\includegraphics[width=18.0cm]{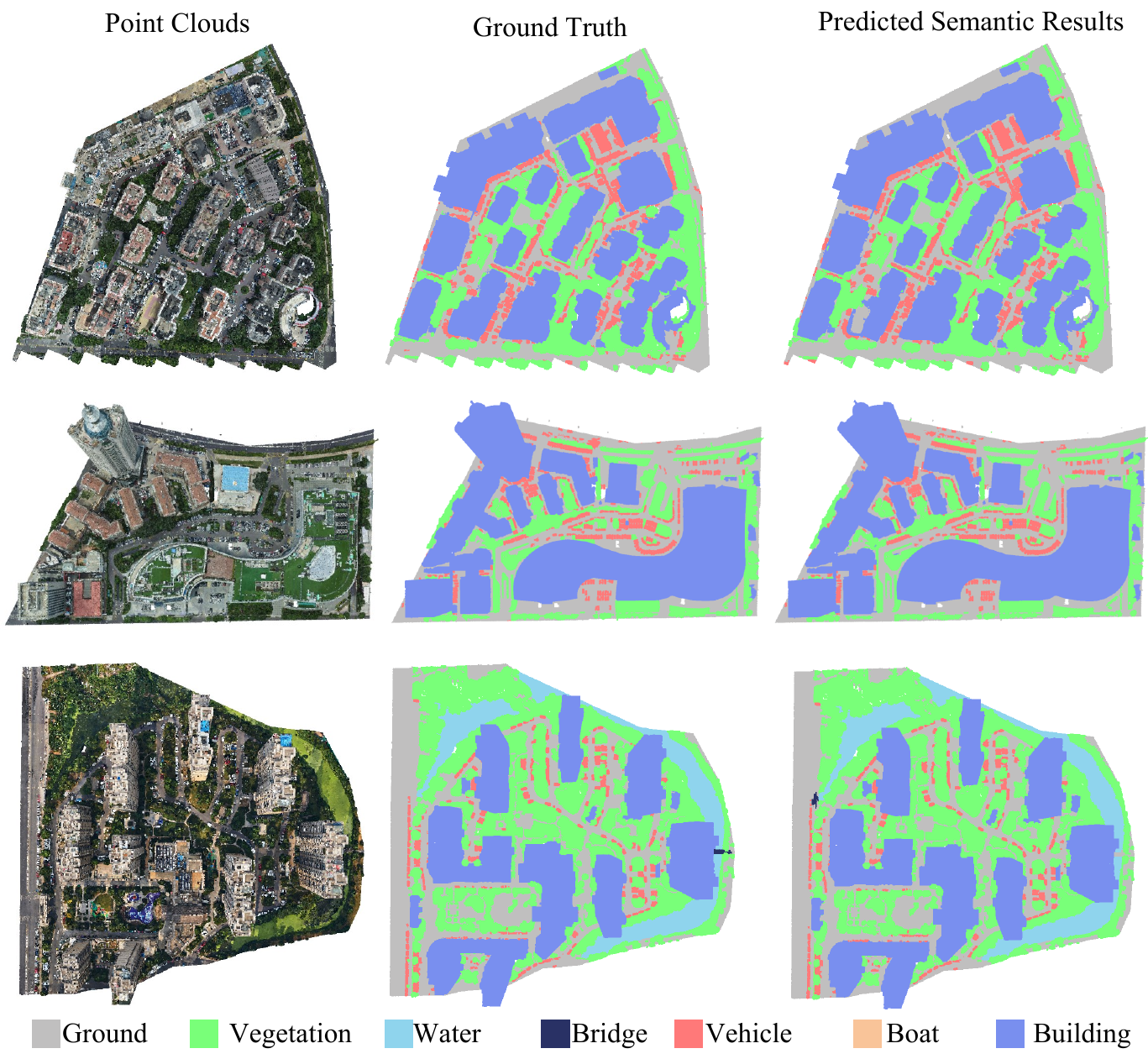}
\caption{{Qualitative results in the open-set benchmark of UrbanBIS dataset~\cite{yang2023urbanbis}.} 
}
\label{fig:urbanbis}
\end{figure*}

\begin{figure*}[t!]
\centering
\includegraphics[width=18.0cm]{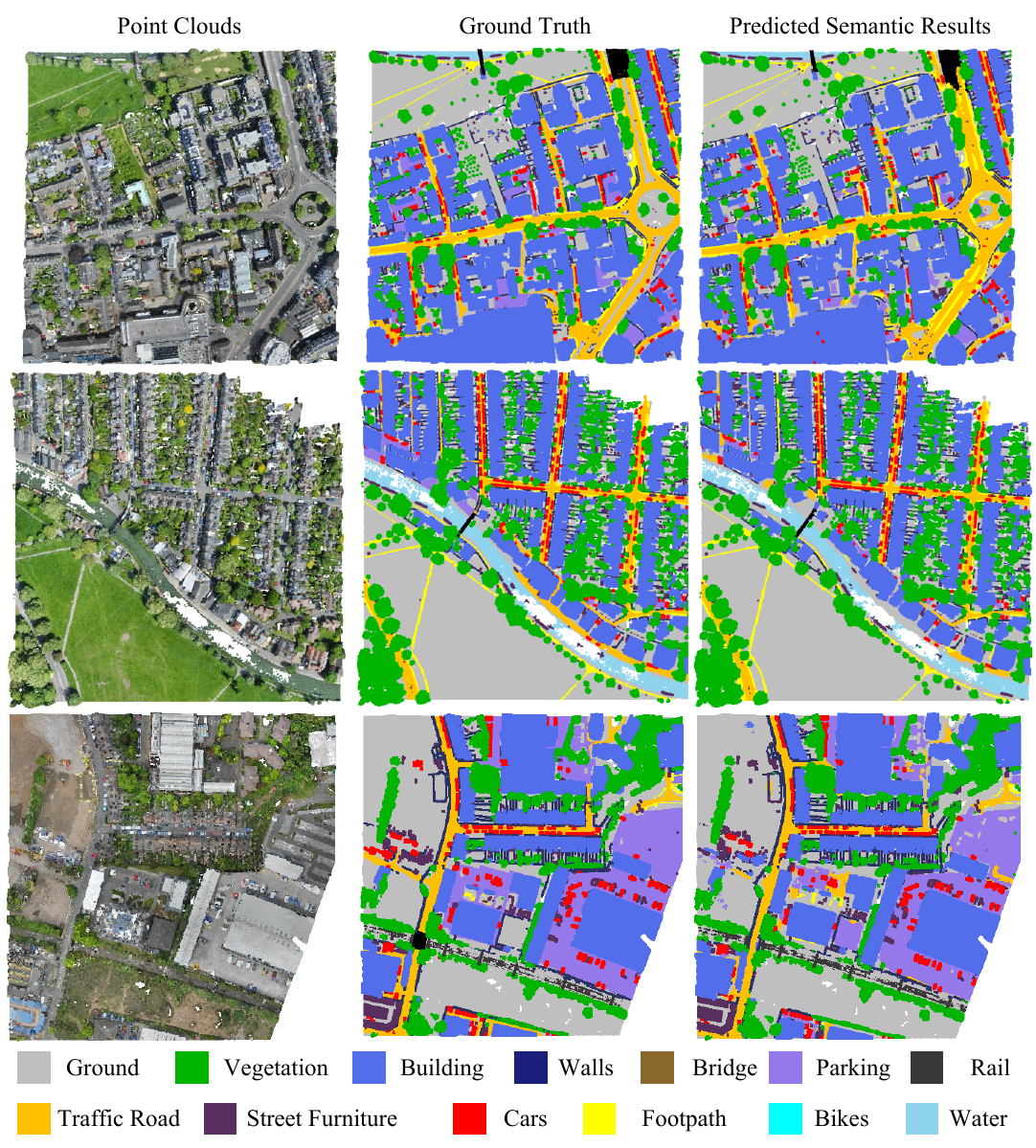}
\caption{{Qualitative results in the closed-set benchmarks of SensatUrban dataset~\cite{hu2022sensaturban}.} 
}
\label{fig:sensaturban}
\end{figure*}

\begin{figure*}[t!]
\centering
\includegraphics[width=18.0cm]{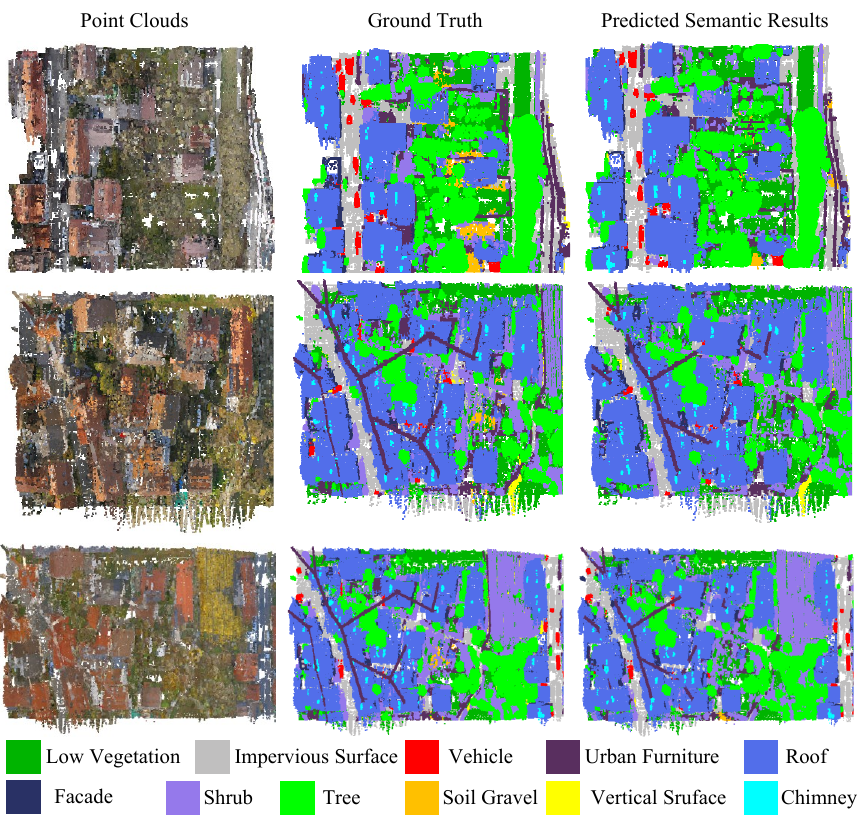}
\caption{{Qualitative results in the closed-set benchmarks of Hessigheim dataset~\cite{kolle2021hessigheim}.} 
}
\label{fig:hessigheim}
\end{figure*}

\begin{figure*}[t!]
\centering
\includegraphics[width=18.0cm]{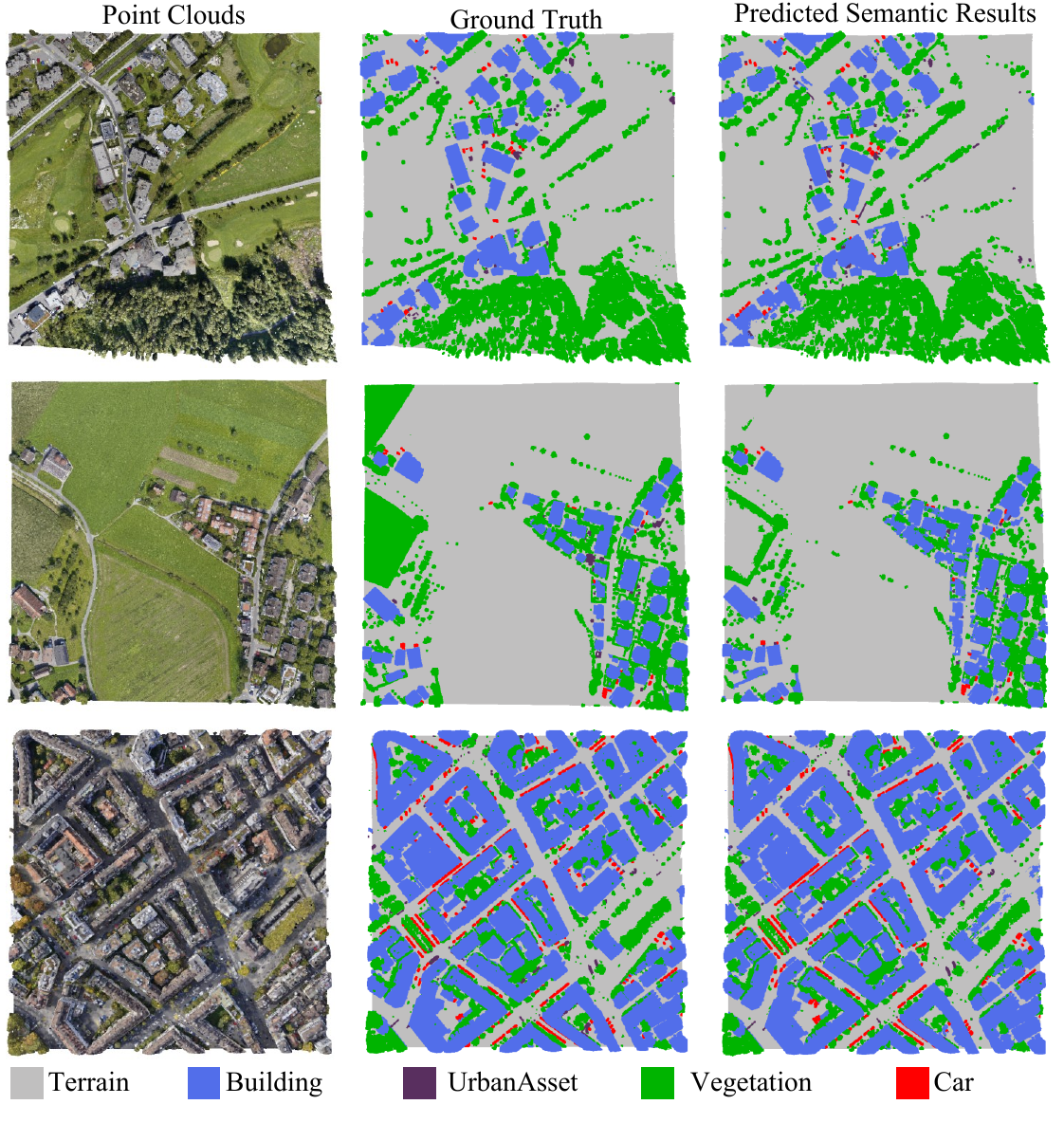}
\caption{{Qualitative results in the closed-set benchmarks of Swiss3DCities dataset~\cite{can2021semantic}.} 
}
\label{fig:swiss3d}
\end{figure*}

\noindent\textbf{Qualitative results.} As shown in Fig.~\ref{fig:urbanbis}, \ref{fig:sensaturban}, \ref{fig:hessigheim}, and \ref{fig:swiss3d}, we present the comparative visualization results of both open-set and closed-set benchmarks. The qualitative analysis demonstrates that our CitySeg achieves precise semantic segmentation of point clouds, with predictions exhibiting remarkable consistency with the ground truth annotations.

\subsection{Ablation Study}
In this subsection, we conduct an ablation study to evaluate the effectiveness of each proposed component and training strategy. All experiments are conducted on the SensatUrban test set using identical hyperparameter settings.

\noindent\textbf{Ablation on Local-global Cross-attention}.
We evaluate the baseline point network's semantic segmentation performance and compare it with the network augmented by the proposed local-global cross-attention module. As shown in Tab.~\ref{tab:ablation}, incorporating the local-global attention module improves OA from 89.8 to 93.6 and mIoU from 65.0 to 67.8.  These results indicate that the global branch in the module provides a valuable reference for fine-grained semantic understanding.
Notably, although the local branch processes up to 65,536 sampled points per batch, far larger than current point backbones' input size, the corresponding 3D receptive field remains inadequate for capturing global context in city-scale scenarios.
Our proposed module effectively leverages sparse global information, thereby enhancing the semantic understanding of city-level point clouds.

\noindent\textbf{Ablation on Hierarchical Classification Strategy.}
As shown in Tab.~\ref{tab:ablation}, incorporating the hierarchical classification method leads to a substantial increase in model performance when trained on multiple datasets: OA improves from 71.4 to 93.6, and mIoU from 39.4 to 67.8. This notable enhancement demonstrates that the hierarchical classification strategy effectively addresses inconsistencies in annotations across diverse data sources. Importantly, these improvements are achieved without altering the network architecture or training procedures, requiring only the assignment of label levels according to the annotation rules of each dataset. The total number of manually annotated samples is the aggregate across categories from all data sources.

\noindent\textbf{Ablation on Hinge Loss.}
We further assess the impact of incorporating the Hinge Loss in our training framework. As shown in Tab.~\ref{tab:ablation}, adding the Hinge Loss increases OA from 91.4 to 93.6 and mIoU from 63.5 to 67.8. These results demonstrate that the Hinge Loss consistently enhances model performance by increasing the feature separation between similar categories, thereby improving fine-grained semantic discrimination.

\section{Conclusion}
In this paper, we present CitySeg, a foundation model for 3D open vocabulary semantic segmentation in city-scale scenarios. When training across multiple datasets, we identify two major challenges: divergent data distributions and non-uniform 3D data annotations. We address these issues by introducing a point network with local-global cross-attention for precise scene understanding, a hierarchical classification strategy to reconcile annotation differences, and a customized two-stage training approach. Extensive experiments demonstrate that CitySeg achieves SOTA performance on nine closed-set benchmarks and one open-set benchmark, significantly outperforming existing methods and offering a robust solution for city-scale 3D semantic segmentation tasks.
\newpage
~\\
~\\
~\\
~\\
~\\
~\\
~\\
~\\
~\\
~\\
~\\
~\\
~\\
~\\
~\\
~\\
~\\
~\\
~\\
~\\
~\\
~\\
~\\
~\\
~\\
~\\
~\\
~\\
~\\
~\\
~\\
~\\
~\\
~\\
~\\
~\\
~\\
~\\
~\\
~\\
~\\
\bibliographystyle{IEEEtran}
\bibliography{reference}

\end{document}